\documentclass[11pt]{article}

\usepackage{acl}
\usepackage[utf8]{inputenc}
\usepackage{CJKutf8}
\usepackage{xcolor}
\usepackage{arabtex}
\usepackage{colortbl}

\usepackage{times}
\usepackage{latexsym}

\usepackage{float}

\usepackage{booktabs}
\usepackage{float,lscape}
\usepackage{array}
\usepackage{makecell}

\usepackage{pgfplots}
\pgfplotsset{width=10cm,compat=1.9}

\usepackage{multirow}

\usepgfplotslibrary{external}
\usepackage[T1]{fontenc}
\usepackage{microtype}

\usepackage[arabic,USenglish]{babel}
\usepackage{inconsolata}
\usepackage{abstract}

\definecolor{lightgreen}{rgb}{0.67, 0.94, 0.82}
\definecolor{darkorchid}{rgb}{0.6, 0.2, 0.8}

\title{In What Languages are  Generative Language Models the Most Formal? Analyzing Formality Distribution across Languages}

\author{As{\i}m Ersoy\textsuperscript{1}\thanks{\hspace{1em}Equal contribution. This work was done as part of the Fatima Fellowship mentoring program.}
*\hspace{.3em}, Gerson Vizcarra\textsuperscript{2,3}*\hspace{.3em}, 
Tasmiah Tahsin Mayeesha \textsuperscript{4}*, Benjamin Muller \textsuperscript{5}   \\
  \textsuperscript{1}Huawei Turkey R\&D Center {\hspace{1em}} 
  \textsuperscript{2}Nisum Latam {\hspace{1em}} \textsuperscript{3}Banco de Crédito e Inversiones {\hspace{1em}}\\ \textsuperscript{4}North South University {\hspace{1em}} 
  \textsuperscript{5}Sorbonne Université \\
  \texttt{asim.ersoy@huawei.com}\hspace{.3em} \texttt{gersonw.vizcarra@gmail.com}\\ \texttt{tasmiah.tahsin@northsouth.edu}\\}
\begin{document}
\maketitle

\begin{abstract}
   
    % \asim{v2}
    % Multilingual generative language models are increasingly fluent in a large variety of languages. Trained in the concatenation of corpora in multiple languages, they enable powerful transfer from high-resource languages to low-resource ones. However, it is still unknown what cultural biases are induced in the predictions of these models. Moreover, in some languages, linguistic style properties such as formality could represent the culture of entire populations.\mayeesha{Furthermore, linguistic style properties such as formality vary greatly across cultures.} Thus, formality-biased models may not be able to adapt properly across different cultures. In this work, we analyze the formality biases of XGLM \citep{lin2021few} and BLOOM \cite{scao2022bloom} across five languages. We conduct the analysis across different dimensions such as the cohesiveness of the generations, the formality bias, and the formality preservation. We find that both models show bias toward formal text when prompted with short neutral prompts. Besides, both models behave differently when preserving the formality level of long prompts. \\

    Multilingual generative language models (LMs) are increasingly fluent in a large variety of languages. Trained on the concatenation of corpora in multiple languages, they enable powerful transfer from high-resource languages to low-resource ones. However, it is still unknown what cultural biases are induced in the predictions of these models. In this work, we focus on one language property highly influenced by culture: formality. We analyze the formality distributions of XGLM and BLOOM's predictions, two popular generative multilingual language models, in 5 languages. We classify 1,200 generations per language as formal, informal, or incohesive and measure the impact of the prompt formality on the predictions. Overall, we observe a diversity of behaviors across the models and languages. For instance, XGLM generates informal text in Arabic and Bengali when conditioned with informal prompts, much more than BLOOM. In addition, even though both models are highly biased toward the formal style when prompted neutrally, we find that the models generate a significant amount of informal predictions even when prompted with formal text. We release with this work 6,000 annotated samples, paving the way for future work on the formality of generative multilingual LMs. 
\end{abstract}

\section{Introduction}

Natural Language Processing (NLP) systems are used worldwide across multiple cultures, audiences, contexts, communication goals, demographics, and languages. Thus it is essential that %the technologies based on
these models be able to adapt to the sociocultural context of its users. %properly.
As described by \citet{hershcovich-etal-2022-challenges}, linguistic style is one of the major dimensions by which cultures vary in NLP technologies. 

% \gerson{double check flow of last two sentences and the claim of the first citation} \asim{The sentence starting with "incorporating.." isn't clear to me at all. I feel like it isn't a natural transition from what is coming before it. I would remove that sentence completely.}
% Incorporating these social factors of language into NLP systems can open new avenues of research and improve the performance of the models in different tasks \cite{hovy-yang-2021-importance}. \mayeesha{this paper has good discussion on cultural awareness needed in language technologies including formality so citing in the next section}

In this work, we focus on formality. Formality is a stylistic property of language that can impact how we perceive a text. It typically carries information about the culture of the speaker (or writer), %audience,  
is constrained by the context of the message, and can impact %describe 
the communicative goal of a text \cite{heylighen1999formality}. Generating text with a desired level of formality can be useful for different NLP applications \cite{hovy-yang-2021-importance}. For example, controlling the tone of machine translation models \cite{sennrich-etal-2016-controlling, niu-etal-2017-study, feely-etal-2019-controlling}, designing chatbots with formality awareness to respond to user-preferred conversational style \cite{10.1145/3543829.3543831}, or assisting users to change the formality level of their writings \cite{rao-tetreault-2018-dear, wang-etal-2019-harnessing, wang-etal-2020-formality}.

Generative language models have demonstrated capabilities in producing cohesive texts and solving NLP tasks with zero/few-shot learning \citep{radford2019language, brown2020language, chowdhery2022palm, zhang2022opt}, even in multilingual scenarios \citep{lin2021few, scao2022bloom, barbieri2022xlm, jiang2022xlm}. Multilingual language models are trained with large amounts of text from different sources. That training process could make the model biased towards a certain level of formality because of the data of each language as well as cross-lingual transfer \citep{pires-etal-2019-multilingual,libovicky-etal-2020-language,muller-etal-2021-first}, limiting the capabilities of the model to adapt to different cultures of an NLP application. %different cultures, audiences, or communication goals of an NLP application. 

This work analyzes the formality level of two multilingual language models: XGLM \citep{lin2021few} and BLOOM \cite{scao2022bloom}, across five languages, namely Arabic, Bengali, English, French, and Spanish. To do so, a native/proficient speaker of each language evaluates the generation outputs of each model into three categories: formal, informal, and incohesive. This evaluation allows us to analyze the generations across three different dimensions: the cohesiveness of the generations,\footnote{In short, we define a sequence as incohesive if it cannot be evaluated as formal/informal. More details in Section \ref{subsec:evaluation_guides}} the formality bias given neutral prompts, and the formality preservation given formal/informal prompts. As an example, we show in Table~\ref{tab:example} the predictions of BLOOM and XGLM conditioned on the same prompt in Bengali but generating text of different formality level. Overall, our contributions are the following:
%\footnote{Incohesivetext is excluded from our formality analysis}

\begin{table*}[t]
\centering
\small
\begin{tabular}{p{3cm}p{5cm}p{5cm}}
     \toprule
     \makecell{\multirow{2.5}{*}{\textbf{Prompt}}}  &  \multicolumn{2}{c}{\textbf{Generation}}\\
     \cmidrule{2-3}
       &  \makecell{\textbf{XGLM(7.5B)}} &  \makecell{\textbf{BLOOM(7.1)}}\\
     \midrule
    \begin{minipage}{\linewidth}
\includegraphics[width=\linewidth]{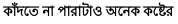}
\\
\\
\\
\\
\\
\\

\end{minipage} \textit{Not being able to cry is also very difficult} &  \begin{minipage}{0.9\linewidth}
\includegraphics[width=\linewidth]{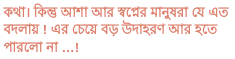}
\\
\\
\\
\\
\end{minipage} \textit{. But people of our hopes and dreams change so much! There could not be a greater example...!} &   \begin{minipage}{\linewidth}
\includegraphics[width=\linewidth]{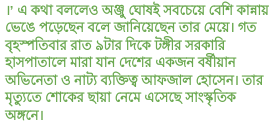}
\end{minipage} \textit{But Anju Ghosh was the one who broke down in tears, her daughter said. Anju Ghosh returned to work after spending time with her daughter. Afzal Hossain, a veteran actor and theater personality of the country, died at Tongi Government Hospital last Thursday around 9 pm.
}
\\
    \bottomrule
\end{tabular}
    \caption{XGLM \cite{Lin2021FewshotLW} and BLOOM \cite{scao2022bloom} generating predictions of different formality (informal in red and formal in green) based on a Bengali formal prompt sampled from the InFormal dataset \cite{krishna-etal-2022-shot}. As illustrated here, the formality of the predictions differ between XGLM and BLOOM. 
    }
    \label{tab:example}
\end{table*}

\begin{itemize}
    \item To the best of our knowledge, this is the first work to analyze the formality of generative multilingual language models across multiple languages. While we have focused on specific models and languages in this work, the procedures followed to define formality, prompt sourcing, language generation, and measurement of how formality is preserved from prompts are generalizable to any generative system and language. We open-source 1,200 generations, per language, manually annotated as formal, informal, or incohesive \footnote{\url{https://github.com/asimokby/formality-bias-analysis}}. 
     
%    \item We find that formal sequences are characterized by having longer sentences in all languages except Arabic. Also, informal generations in English, French, and Spanish are characterized by including more dialogue marks, and in Bengali, by having more punctuation marks. 

 %   \asim{the following point is an updated version of the prev bullet point}
    
    \item We find that BLOOM generates about twice longer texts as XGLM. Besides, almost all the generated formal sentences are longer than the informal ones. Also, informal generations in English, French, and Spanish are characterized by being more conversational, and in Bengali, by having more punctuation marks.

    \item We find that BLOOM is significantly more cohesive than XGLM in English, French, and Spanish and performs similarly in other languages. 

    %\item Both XGLM and BLOOM are generally biased toward formal text when prompted in a neutral way. However, both models are very sensitive to an informal prompt and will generate informal text if prompted this way. This is particularly striking for Arabic: BLOOM is able to preserve dialectal Arabic (considered informal) while being mostly pretrained on Modern-Standard Arabic. 

   % \asim{the following point is an updated version of the prev bullet point}

    \item Both XGLM and BLOOM are generally biased toward formal text when prompted in a neutral way. However, both models are very sensitive to the formality of the prompt and will generate informal text if conditioned with informal prompt. This is particularly striking for Arabic: BLOOM generates dialectal Arabic (considered informal) when prompted with informal text while being extremely biased toward Modern-Standard Arabic (considered formal). 
    
\end{itemize}

\section{Formality Across Different Languages}\label{sec:formality}

We start by defining formality in the five languages of our study. 

\paragraph{Arabic} The Arabic language is spoken in many dialects \citep{janet1}. %and, therefore, the region's name is used as an identifier for each (for example, Egyptian Arabic). \ben{removed cause I dont think it is necessary}
These dialects are variants of classical or standard Arabic, which has a modernized version of it called Modern Standard Arabic (MSA). \citet{badawi1973mustawayat}, in his famous book \textit{Mustawayat Al-arabiyya Al-muasira Fi Misr (The levels of contemporary Arabic in Egypt)}, presents a theory on the relationship between standard Arabic (\textit{Fusha}) and vernacular Arabic (\textit{Ammiya}) in Egypt. His theory describes the situation as a continuum with 5 major divisions: illiterate colloquial Arabic, educated colloquial Arabic, elevated colloquial Arabic, modern standard Arabic, and classical Arabic. The first three divisions are \textit{Ammiya}, which is considered informal and not necessarily grammatically correct. The last two divisions are \textit{Fusha}, which is considered formal. However, the definition of what is formal and what is informal could depend on the problem at hand, for example, in one case, elevated colloquial Arabic could be considered formal while illiterate colloquial Arabic as informal. In our work, we define formality for Arabic as follows: a piece of text is formal if it contains no words coming from any Arabic dialect which is not considered as \textit{Fusha}, following \cite{badawi1973mustawayat}'s definition of \textit{Fusha}. For example, the following sentence: \AR{أين أقرب مسجد؟
} 
\textit{where is the closest mosque?} is formed of only \textit{Fasih}, formal, words. Similarly, a piece of text is informal if it contains a word coming from any dialect and not \textit{Fusha}. For example, \AR{فين أقرب مسجد؟ } \textit{where is the closest mosque?} is informal because of the word \AR{فين} \textit{where} which is coming from Egyptian Arabic. 

\paragraph{Bengali}
Bengali has a complex and elaborate system of using pronouns to express the degrees of familiarity and formality between the participants in a conversation \cite{das_1968,uddin2019second}. T-V distinction \cite{brown1960pronouns} or the contextual usage of pronouns to convey varying levels of formality, familiarity, and politeness, which is found in many Romance languages (French, German, Italian, Spanish, etc), can also be seen in Bengali. Bengali follows a tripartite form of second-person pronouns like other South Asian languages, including Hindi/Urdu \cite{bhatt2012honorifics,bhatt2015acquisition}, Malaysian \cite{mcginn1991pronouns} and can be considered a T/N/V language \cite{thompson_2006,uddin2019second} with an added level of neutral or semi-formal tone aside from formal and informal. 

The set of pronouns to be used depends on the relationship between the speaker and the audience and the intimacy level. For instance, \textit{you} in English has three different variations in Bengali,   \begin{minipage}{.09\textwidth}
\includegraphics[width=\linewidth]{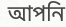}
\end{minipage} / Apni (formal)
for respected elders and strangers, 
\begin{minipage}{.05\textwidth}
\includegraphics[width=\linewidth]{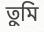}
\end{minipage}
/ Tumi (polite) for siblings/friends or familiar people and
\begin{minipage}{.05\textwidth}
\includegraphics[width=\linewidth]{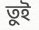}
\end{minipage} / Tui (informal) for those who are younger, children or very close friends. The third person \textit{he / she} can be translated to \begin{minipage}{.07\textwidth}
\includegraphics[width=\linewidth]{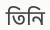}
\end{minipage}/Tini (formal) vs 
\begin{minipage}{.05\textwidth}
\includegraphics[width=\linewidth]{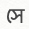}
\end{minipage}
/Se (informal) which encodes two levels of formality- honorific and non-honorific. Bengali Pronouns can encode numbers such as singular/plural, but the notion of formality is not changed by gender or numerical properties  \cite{David+2015}. 

 The following are other considerations of formality in Bengali :  

\begin{itemize}
\item Texts containing a high frequency of Sanskrit-originated words can be considered formal. Agglutination/Compound words can be considered more formal compared to their analytical or elaborated forms. 
\begin{minipage}{.09\textwidth}
\includegraphics[width=\linewidth]{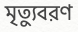}
\end{minipage}  
(formal) /
\begin{minipage}{.1\textwidth}
\includegraphics[width=\linewidth]{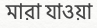}
\end{minipage} 
(informal) ---  \textit{death} has same meaning, but a different formality \cite{panda1992rule,nagarajan2014constraints,ghosh2022bangla}.

\item Bengali pronouns agree with the verb in levels of formality and there are formal and informal variations of the same verb. %\gerson{I think it is only necessary to mention the second idea} %\mayeesha{No, I think its important to mention that verb forms are switching depending on the pronoun usage}. 
\cite{David+2015,Sultana2016DescriptionOV} For instance, verbs like \textit{Give}, \textit{Eat}, \textit{Go} can be written as 
\begin{minipage}{.14\textwidth}
\includegraphics[width=\linewidth]{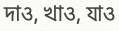}
\end{minipage}
(formal) or 
\begin{minipage}{.10\textwidth}
\includegraphics[width=\linewidth]{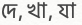}
\end{minipage} 
(informal) depending on the context. 
\item Bengali does not contain any negative pronoun or adverb and sentences can be modified to be negative at a syntactic level by adding 
\begin{minipage}{.05\textwidth}
\includegraphics[width=\linewidth]{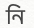}
\end{minipage} or other modifiers. These negation modifiers like \textit{na/nei/nai/Ni} can indicate variations in formality \cite{thompson_2006}.
\item Among Bengali speakers in Bangladesh, regional dialects like Sylheti, Chakma, and Chittagonian are generally considered deviant or informal while classical Bengali dialect (Sādhubhāsā) or standardized Bengali dialect (Cholito vasha) is considered formal \cite{ray1966bengali}.

\end{itemize}

\paragraph{English}
Formality in English is commonly defined as the style of language used in a given situation. A formal speech, for instance, has a very careful selection of pronunciation, words, and structure \citep{richards2013longman}. \citet{heylighen1999formality} divide English formality into two dimensions: a \textit{deep formality}, characterized by the understanding of the precise meaning, avoiding ambiguity; and a \textit{surface formality} which focuses on the rigorous selection of manners. Some recent works focus on the latter to evaluate formality using the selection of words \citep{brooke2010automatic} and discarding the topic \citep{pavlick-tetreault-2016-empirical}. In accordance with \citet{liardet2019defining}, we use the following rules to evaluate cohesive English text as informal:

\begin{itemize}
    \item Presence of contractions, abbreviations, and colloquial expressions.
    \item Presence of grammar infelicities, that is, unsuitable expressions, inconsistencies in writing, and misspellings.
    \item High occurrence of delexical verbs and phrasal verbs.
    \item Higher involvement of human participants and subjective judgments, such as opinions.
\end{itemize}

\paragraph{French}
Formality is typically classified in French into three classes: \textit{soutenu}, \textit{courant} and \textit{familier} \cite{gadet:halshs-00114889,beeching2009sociolinguistic}. The register \textit{soutenu} is reserved for legal documents, literature, or when addressing someone we want to show particular respect (e.g., a judge). It usually involves addressing someone with the second singular person (called \textit{vousvoîment}). The register \textit{courant} corresponds to the one used in day-to-day life, for instance when we talk to someone new which is typically neutral and includes few grammatical errors. The register \textit{familier}
is the one used with friends, or within a family circle. It usually involves addressing someone with the second singular person tu (\textit{tutoîment}). It can include a large portion of grammatical errors. It can also include slang and insults in their most vulgar form.
In this work, following what was done in the XFORMAL work \citep{briakou-etal-2021-ola}, we classify generated text into two classes. Soutenu is associated with the formal class while \textit{familier} and \textit{courant} with the informal class. 

\paragraph{Spanish} Formality in Spanish is commonly described by the T-V distinctions in the singular second-person pronoun derived from Latin. Specifically, there are two possible translations for the English pronoun "\textbf{you}": \textbf{\textit{tú}} is considered informal while \textbf{\textit{usted}} is formal. Both pronouns have different conjugations. Thus, the formality in sentences that use the singular second person is easily recognizable. 

In the case of the other pronouns, the first person is often considered less polite than the third one \cite{stewart2001pronouns}. For that reason, the third person is commonly used in scientific texts \cite{salazar2013cross}. Aside from the pronouns and their conjugations, according to \citet{cepeda2007redaccion}, a formal text in Spanish should accomplish other characteristics such as:
\begin{itemize}
    \item Having no typographical or grammatical errors. 
    \item Being a set of sentences referring to the same topic.
    \item Being arranged in paragraphs and having a coherent correlation between ideas using appropriate connectors.
\end{itemize}

In our work, we check the presence of slang or offensive terms in a sequence to classify text as informal. Then, T/V distinction in sentences written using the second person defines the formality level. In a similar way, sentences written in the third person have a bigger probability of being classified as formal compared to the ones written in the first person. The final priority is the layout: paragraph-structured sequences are considered as formal in more scenarios than conversational-structured ones.

\section{Related Work}

\paragraph{Biases of Generative Language Models}

Recent literature on Large Language Models (LLMs) demonstrated social bias and prejudice against minorities \cite{sheng-etal-2021-societal,blodgett-etal-2020-language,10.1145/3442188.3445922,bommasani2021opportunities,Liang2021TowardsUA} in terms of many categories including gender \cite{sun-etal-2019-mitigating,cao-daume-iii-2020-toward,Felkner2022TowardsWD}, race \cite{davidson-etal-2019-racial}, religion \cite{10.1145/3461702.3462624,malik-etal-2022-socially}, occupation, politics and disabilities which result in the production of damaging content. Evaluating social bias and harm produced by monolingual language models is hard, but difficulties increase in multilingual settings. To create multilingual evaluation frameworks, it has been argued that careful curation of culturally aware datasets and knowledge of cultural differences that exist between languages is necessary \cite{Talat12022YouRW}. %along with increasing transparency in the documentation and diversifying the stereotypes that are investigated \ben{removed cause not necessary}

Many papers have focused on measuring social biases and stereotypes against historically disadvantaged groups and counteracting them for a limited number of languages like English \cite{nadeem-etal-2021-stereoset, nangia-etal-2020-crows, barikeri-etal-2021-redditbias}, French \cite{neveol:hal-03629677}, Hindi \cite{malik-etal-2022-socially}, but similar work has not been done for low-resource languages like Bengali. Since LLMs such as BLOOM \cite{scao2022bloom} can be continuously (re)trained and are deployed by companies to be accessible by users, proposals have been made to create social bias verification pipelines for LLMs similar to software testing \cite{nozza-etal-2022-pipelines}. To our knowledge, the evaluation of multilingual models for measuring cultural biases like formality has not been attempted so far. 

\paragraph{Formality Analysis}

Previous work in formality analysis has focused on formality classification \cite{heylighen1999formality,5587767,pavlick-tetreault-2016-empirical,dementieva2022detecting}, formality style transfer in English \cite{rao-tetreault-2018-dear,wang-etal-2019-harnessing,wang-etal-2020-formality,czeresnia-etinger-black-2019-formality,madaan-etal-2020-politeness,yao-yu-2021-improving,briakou-etal-2021-evaluating}, and in the multilingual setting \cite{korotkova2019grammatical,briakou-etal-2021-ola, krishna-etal-2022-shot}. Formality-sensitive machine translation to control the generation of machine translation models to target formality has received attention in recent years \cite{sennrich-etal-2016-controlling,niu-etal-2017-study,feely-etal-2019-controlling,10.1007/978-3-030-46140-9_29,niu2020controlling,schioppa-etal-2021-controlling} and benchmark MT datasets and models have been published \cite{nadejde-etal-2022-cocoa,rippeth2022controlling}.

Recently, several datasets with formality annotations have been introduced in multiple languages. Initial attempts included annotating sentences from various resources such as emails, news, online forums, and blog sentences with numerical formality rating \cite{lahiri2015squinky,pavlick-tetreault-2016-empirical}. The Grammarly’s Yahoo Answers Formality Corpus (GYAFC) \cite{rao-tetreault-2018-dear} is a benchmark formality style transfer dataset for English. XFORMAL \cite{briakou-etal-2021-ola} extended formality style transfer to the multilingual setting by collecting data for four European languages (Brazilian, Portuguese, French, and Italian). InFormal (Indic Formality Evaluation Dataset) \cite{krishna-etal-2022-shot} is a small dataset of 4k samples in four Indic languages - Hindi, Bengali, Kannada, Telugu with crowdsourced formality annotations. TAOCD (The Arabic Online Commentary Dataset) \cite{zaidan-callison-burch-2011-arabic} presents an annotated dataset of informal Arabic with high dialectal content with 108k labeled sentences. In our work, we use GYAFC (English), XFORMAL (French), TAOCD (Arabic), and InFormal (Bengali) to source prompts for our analysis of language models along with other resources described in table \ref{tab:prompt_details}. In the following sections, we describe our experiments and results for different languages. 

\section{Experiments}
 % \asim{I think we also evaluate other things. We could mention all the dimensions here} 
We evaluate different dimensions of formality of the generation outputs of two state-of-the-art generative multilingual language models: XGLM 
 \citep{lin2021few} and BLOOM \citep{scao2022bloom}, in five languages: Arabic, Bengali, English, Spanish, and French. We hypothesize that the influence of high-resource languages in the corpus can involve biases in the formality of the whole models. To see their behavior in different scenarios, we employ distinct variations of prompt lengths and formality. In addition, we tweak some parameters when generating to avoid incohesive outputs.
 
\subsection{Language Models}

\paragraph{XGLM} \citep{lin2021few} is a multilingual generative language model based on a decoder transformer. XGLM is trained with 500 billion tokens belonging to 30 languages. XGLM aims to achieve multilingual zero-shot and few-shot learning performance for different tasks. To do so, their authors propose multilingual prompting to improve the results of single-language prompts. XGLM has five sizes according to their number of parameters ranging from 564 million to 7.5 billion parameters. %: from 564 M, 1.7 B, 2.9 B, 4.5 B, and 7.5 B. 
We employ the models with 2.9 and 7.5 billion parameters for this study\footnote{We use the checkpoints and implementations from \url{https://huggingface.co/models}}.

\paragraph{BLOOM} \citep{scao2022bloom} is also a multilingual generative language model trained on around 341 billion tokens from a corpus of 59 languages (13 of them are programming ones) to democratize huge pre-trained language models. BLOOM was trained from a collection of multiple sources such as Huggingface datasets, Github code, and Web Common Crawl. The data sources were then preprocessed to reduce non-natural language and anonymize personal identifiable information. BLOOM used architectural improvement introduced with the Megatron-LM GPT2 \citep{shoeybi2019megatron},   %Among the variations in 
such as a normalization layer after the embeddings, ALiBi positional embeddings \citep{press2021train}, and a Byte-Level Byte Pair Encoding \citep{radford2019language}. BLOOM was released in different sizes ranging from 560 million %, 1.1 B, 3 B, 7.1 B, 
to 176 billion parameters. We use the 3B and 7.1B parameter checkpoints\footnotemark[2] for our experiments as they can be compared to XGLM ones.

XLGM and BLOOM are decoder-based transformers pre-trained on a similar set of languages with a comparable amount of data. We compare checkpoints of similar scale (i.e. we compare XGLM 2.9B with BLOOM 3B and XGLM 7.5B and BLOOM 7.1B). Regarding the proportion and data sources on which both models were trained, BLOOM was trained on a more varied set of domains than XGLM in spite of the XGLM corpus being larger. In addition, the BLOOM corpus has a more balanced distribution of the amount of data of the languages evaluated in this study. More details about the quantity and sources of both models can be found in Appendix \ref{sec:appendix1}.

\subsection{Prompting for Formality Evaluation} 

We employ two prompting strategies to condition the generation of the models. In that way, the behavior of the model in different scenarios can be assessed. % properly. %\mayeesha{ to assess the behavior of the models in different scenarios - merge two sentences together? }

\paragraph{Short Neutral Prompts}
A short prompt is composed of up to three words to condition the language of the output without giving any context that could impact the formality level. That allows us to measure the models' tendency to produce a certain formality level with a neutral input. For the lexicon of each language\footnote{\url{http://corpus.rae.es/lfrecuencias.html},\\\url{https://www.pinhok.com/kb/bengali/98/100-basic-bengali-vocabularies/},\\\url{https://talkinarabic.com/arabic-words/},\\\url{https://en.wikipedia.org/wiki/Most_common_words_in_English}\\\url{https://strommeninc.com/1000-most-common-french-words-frequency-vocabulary/}}, we pick a set of common words (or a combination of them to avoid the confusion of languages when generating) that can be used in both formal and informal sentences. 
We illustrate the short prompt we use in Table~\ref{tab:prompt_details}.

\paragraph{Long Informal/Formal Prompts}
% \ben{this should describe the origin of the informal/formal promtps and point to the table too here}
This set of prompts is composed of truncated sentences extracted from existing formal/informal sources. Using these prompts, we can verify how much the models preserve the formality level of their input. The sources of the prompts include formality datasets such as GYAFC \citep{rao-tetreault-2018-dear}, XFORMAL \citep{briakou-etal-2021-ola}, InFormal \citep{krishna-etal-2022-shot}. We also include dataset crawlings from webs \citep{zaidan-callison-burch-2011-arabic, canete2019compilation} and informal songs \citep{munoz2018rap}. 

% \asim{We could talk here about the Arabic dataset as well} \gerson{Isn't Zaidan (2011)? An option is to add some explanations of each dataset, but I think it shouldn't be our focus} \mayeesha{how about pointing directly to the table for the prompt sources and removing the dataset citations, because they have been added in the related work section too. We should say what we mean by preserving the formality level of the input-->generating sentences with same formality}

Table \ref{tab:prompt_details} details which words/group of words we use as short prompts and the dataset sources of the formal/informal prompts for each language.

% \\
% We evaluated the model's generation outputs using three prompting strategies for each assessed language: (a) a very short prompt that represents a neutral input, (b) a formal prompt, and (c) an informal prompt. Two key concepts in using different prompt types are measuring the model's tendency to generate a certain formality level with a neutral input (i.e. a short prompt which is neither formal nor informal \ben{added}) (a) and verifying how much a multilingual language model preserves the formality level of its input (b, c). \ben{this last sentences: 'Two key concepts' is hard to understand, rephrase}

% We employ the most common words (or the combination of them to avoid the confusion of languages when generating) of each language that appear of sentences \citep{rae2008frecuencias} \gerson{add references for other languages} as short prompts. Table \ref{tab:prompt_details} shows the specific details of the prompt sources of all evaluated languages. 
% \ben{We pick common words from existing lexicon\footnote{add url footnotes} that can be both used in the formal and informal sentences.}
% \ben{recall how we got this most frequent words by citing previous work or naming the datasets used}. Also, the formal and informal prompts were extracted from different sources for each language. 

\begin{table}[t]
\footnotesize
\centering
\begin{tabular}{c p{2.8cm}p{1.4cm}p{1.4cm} }
 \toprule
  & \makecell{\textbf{Neutral$^+$}} & \makecell{\textbf{Formal*}} & \makecell{\textbf{Informal*}}\\
  \midrule
 \textbf{ar} & \AR{لما} \textit{(When/Then)}, \AR{نعم} \textit{(Yes)}, \AR{هناك} \textit{(There)}, \AR{لولا} \textit{(Unless)}, \AR{لو} \textit{(If)}, \AR{من} \textit{(From)}, \AR{عند} \textit{(At/When)}, \AR{والله} \textit{(I swear)}, \AR{في} \textit{(In)}, \AR{لا} \textit{(No)} & TAOCD \citep{zaidan-callison-burch-2011-arabic} & TAOCD \citep{zaidan-callison-burch-2011-arabic} \\
 \textbf{bn} & 
\begin{minipage}{.04\textwidth}
\includegraphics[width=\linewidth]{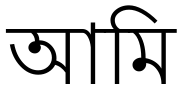}
\end{minipage} \textit{(I)},
\begin{minipage}{.035\textwidth}
\includegraphics[width=\linewidth]{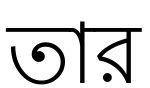}
\end{minipage} \textit{(His/Her)},
\begin{minipage}{.025\textwidth}
\includegraphics[width=\linewidth]{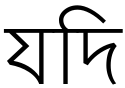}
\end{minipage} \textit{(If)},
\begin{minipage}{.03\textwidth}
\includegraphics[width=\linewidth]{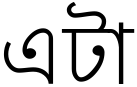}
\end{minipage} \textit{(It)},
\begin{minipage}{.02\textwidth}
\includegraphics[width=\linewidth]{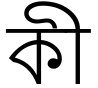}
\end{minipage} \textit{(What)},
\begin{minipage}{.035\textwidth}
\includegraphics[width=\linewidth]{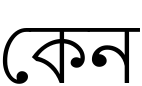}
\end{minipage} \textit{(Why)},
\begin{minipage}{.02\textwidth}
\includegraphics[width=\linewidth]{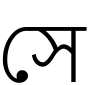}
\end{minipage} \textit{(He/She)},
\begin{minipage}{.05\textwidth}
\includegraphics[width=\linewidth]{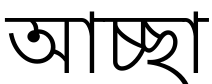}
\end{minipage} \textit{(OK)},
\begin{minipage}{.037\textwidth}
\includegraphics[width=\linewidth]{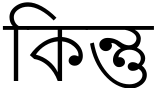}
\end{minipage} \textit{(But)},
\begin{minipage}{.04\textwidth}
\includegraphics[width=\linewidth]{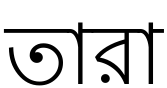}
\end{minipage} \textit{(They)}
 & InFormal \citep{krishna-etal-2022-shot} & InFormal + Microblog dataset \citep{chowdhury2014performing}\\
 \textbf{en} & \textit{The, I, This, He, She, You, They, We, Do, There} & GYAFC \citep{rao-tetreault-2018-dear} & GYAFC \citep{rao-tetreault-2018-dear} \\ 
 \textbf{fr} & C'est \textit{(It is)}, Ils \textit{(They)}, Elles \textit{(They)}, Il \textit{(He)}, elle \textit{(She)}, ce \textit{(This)}, Est-ce que \textit{(\textit{question})}, Ça \textit{(That)}, Ce \textit{(This)}, Deux \textit{(Two)} & XFORMAL \citep{briakou-etal-2021-ola} & XFORMAL \citep{briakou-etal-2021-ola}\\
 \textbf{es} & Por la \textit{(For the)}, Las \textit{(The)}, Los \textit{(The)}, Por el \textit{(For the)}, Con unos \textit{(With some)}, Por que la \textit{(Why the)}, Se ha \textit{(It had)}, Por su \textit{(Because of)}, Para un \textit{(For a)}, De una \textit{(Of a)} & Wikipedia \citep{canete2019compilation} & 9322 rap lyrics in Spanish (filtered) \citep{munoz2018rap}\\ 
 \bottomrule
\end{tabular}
\caption{Prompts used in our experiments. $^+$List of the short prompts across the 5 languages. 10 prompts per language are used with 10 generation sampled for each prompt. *Sources of the formal/informal prompts. 100 prompts per language are sampled from these datasets.}
\label{tab:prompt_details}
\end{table}

\subsection{Generation Parameters}
%  Our generation process includes a set of parameters that enhance \ben{'enhance' is confusing: rephrase in a more simple way: we only selected the parameters that produce the most fluent text} the model
Decoding parameters are important because they can affect the output of a language model directly. For each language, we select a set of parameters to produce fluent text that can be evaluated properly. All selections were chosen to impact the natural formality level of models as less as possible. This subsection presents our list of generation parameters to reproduce our experiments.
% We use several parameters to generate sequences that can be evaluated. Some of them are global, e.g., the minimum length of generation parameter. Others are language specific, e.g., the temperature of token probabilities.

% \ben{try to give some intuition on why we use specific parameters before digging into details in 4.3.1: \\
% - we want the experiments to be reproducible\\
% - we want to use the language model so that it generates real text of a significant length so that we can assess the formality \\
% - we don't want to impact the formality based on our parameters
% }
% \mayeesha{ is there any parameter which we avoided because it could have affected formality?may be that can be mentioned }
\paragraph{Global generation parameters}
Our evaluation of the models is based on the formality of the outputs of each model. Very short sentences, code snippets, and outputs in other languages cannot be evaluated properly. This set of parameters is a collection of language-independent configurations to produce an assessable amount of outputs with a significant length to be evaluated.
\begin{enumerate}
    \item We filter out the generation sequences that are not natural language (i.e., code) by excluding from the generation process all the tokens that contain any of the following symbols: $\{$, $\}$, $($, $)$, $[$, $]$, $\setminus\setminus$, $<$, $>$, $|$, and $;\setminus n$. 
    % the \textit{bad\_words\_ids}\ben{rephrase: most people don't know what function you are talking about. For instance: 'We force the model not to generate code by rejecting code-specific tokens during generation' as a footnote you can point to the function and the bad\_words argument} parameter and including all tokens that contain the following symbols: $\{$, $\}$, $($, $)$, $[$, $]$, $\setminus\setminus$, $<$, $>$, $|$, and $;\setminus n$.
    \item We force the model to generate at least 30 new subword tokens (excluding the prompt) to have a long enough generation sequence and be able to assess formality.
    \item We set a maximum of 150 new tokens of generation to avoid long outputs that could include multiple formality variations. %\ben{TODO: me} %\gerson{Double-check this, ask Benjamin}
    % \ben{same comment: describe what we do, not the code we use to generate the text}
    \item Length of the prompts. For the short-prompt setting, we employ at most three tokens to condition the generation in the desired language. For the formal/informal prompts, we use 15 words (tokenization with white spaces) on average.
    
%Number of generated sequences per prompt: For the short-prompt experiment, we sampled ten sequences per prompt.
%    For the formal/informal settings, we generated one sequence per prompt as the larger prompts constraints much more the topic of the output. 
\end{enumerate}    
Regarding the total number of evaluated outputs, we generated three sets for each evaluated model and language: 100 with short, 100 with formal, and 100 with informal prompts. That resulted in 1200 generated outputs for each language.

\paragraph{Language-specific generation parameters}
Before generating the sequences for formality evaluation, we tweaked some logit parameters for each language. All modifications were done to obtain more fluent sequences and reduce incohesive outputs such as ones with generation repetitions or non-understandable text. This process was done with a varied set of prompts regardless of length and formality level. 

We use sampling to obtain the generation outputs for both models. Three specific parameters were set for both models: We set \textbf{top-k} to 50, which truncates the number of tokens to sample from. We set a high \textbf{top-p} \citep{holtzman2019curious} to generate diverse sampled tokens by cumulative frequency, and a high \textbf{temperature} \citep{ackley1985learning}, which does not skew the distribution towards high probability tokens. The specific details of the parameters can be found in Appendix \ref{tab:gen_parameters}.

\subsection{Formality Evaluation}
\label{subsec:evaluation_guides}
% \ben{Present the categories: formality, informal, NA and data collection process}
% \textbf{We annotate the generation without looking at the prompt!}

We assessed the formality of all generated outputs. To do so, one native/proficient speaker of each language classified all 1200 generated sequences individually. We opted for this evaluation procedure because, at the time of performing the experiments, to our knowledge, there were no multilingual formality classifier models that include Arabic, Bengali, English, Spanish, and French. To avoid possible biases, each generated output was annotated without looking at its prompt and in a randomized order.

The classification categories for all languages are \textbf{formal}, \textbf{informal}, and \textbf{incohesive}. A sequence is classified as formal or informal according to the rules of each language described in section \ref{sec:formality}. The "Incohesive" label is only assigned under certain conditions, such as sequences written in other languages, non-understandable text, very short sequences that cannot be evaluated for formality level, or code snippets.

% However, we asked the evaluators to indicate the priorities when deciding to annotate any sequence as formal, informal, or non-assessable.
% \begin{itemize}
%     \item The evaluation in Bengali was primarily guided by \gerson{Mayeesha, can you write some specific and brief details of the evaluation priorities in Bengali?}
%     \item For Spanish, the priority was to check the presence of slang or offensive terms in a sequence to classify it as informal. Then, T/V distinction in sentences written using the second person guided the classification. In a similar way, sentences written in the third person had a bigger probability of being classified as formal compared to the ones written in the first person. The final priority was the layout: paragraph-structured sequences were classified as formal in more scenarios than conversational-structured ones. Finally, the non-assessable classification was given to sequences written in other languages, non-understandable text, and short sequences that do not contain a clear formality level.
%     \item The assessment in Arabic followed \gerson{Asim, can you write some specific and brief details of the evaluation priorities in Arabic?}
% \end{itemize}

\section{Results \& Analysis} \label{label:results_section}

% \subsection{Generation Quality}

% \ben{moved to fluency}
% The difference in the quality of the generations is seen in the cohesiveness, grammar, and spelling of the generated texts. [maybe elaborate on this and show examples here?]

% The quality of the generations varies from one language to another. One possible reason is the fact that not all the languages are present in the data that the data was trained on equally. Another reason could be the choice of the generation parameters. For example, we notice the following differences between the Arabic and English generations:....[to be continued]

% \subsection{Formality Analysis}

We interpret our results across different dimensions.
We start by analyzing the cohesiveness of each model. We then exclude the incohesive text from our formality analysis. 
% \begin{table}
% \resizebox{\columnwidth}{!}{
% \begin{tabular}{|c| c c c c c|} 
%  \toprule
%  Model/Language & Arabic & Bengali & English & French & Spanish \\ [0.5ex] 
%  \hline
%  BLOOM(7.1B) & \textbf{19\%} & 1\%  & \textbf{8\%}  & 9\%  & 2\%  \\ 
%  BLOOM(3B) & 15\% & \textbf{4\%}  & 2\%  & \textbf{13\%}  & \textbf{3\%}  \\
%  \hline
% XGLM(7.5B) & \textbf{4\%} & 3\%  & \textbf{24\%}  & \textbf{21\%}  & \textbf{17\%}  \\
% XGLM(2.9B) & 3\% & \textbf{4\%}  & 16\%  & 18\%  & 15\%  \\
% \bottomrule
% \end{tabular}
% }
% \caption{Percentages of incohesive texts generated by each model for every language when given a neutral prompt. 
% }
% \label{table:4}
% \end{table}

\begin{table}
\resizebox{\columnwidth}{!}{
\begin{tabular}{c c c c c c} 
 \toprule
 \textbf{Model/Language} & \textbf{Arabic} & \textbf{Bengali} & \textbf{English} & \textbf{French} & \textbf{Spanish} \\ [0.5ex] 
 \midrule
 XGLM(2.9B) & 9.3\% & 8.0\% & 6.7\% & 16.0\% & 6.7\% \\
 BLOOM(3B) & 13.3\% & 4.3\% & 3.3\% & 12.0\% & 3.3\% \\
 \midrule
XGLM(7.5B) & 8.7\% & 5.0\% & 10.0\% & 18.0\% & 7.7\% \\
BLOOM(7.1B) & 12.3\% & 6.3\% & \textbf{3.7\%*} & \textbf{8.7\%*} & \textbf{2.7\%*} \\
\bottomrule
\end{tabular}
}
\caption{Percentages of the incohesive samples out of the 1200 generated samples per language (300 samples per model). Percentages are averaged across prompt types: 400 neutral, 400 formal, and 400 informal prompts. Bolded values show that the corresponding model is significantly better according to a permutation-based statistical test with a p-value of 5\% or less.}
\label{table:incohesiveness_percentages}
\end{table}

% \begin{table}
% \resizebox{\columnwidth}{!}{
% \begin{tabular}{c c c c c c} 
%  \toprule
%  Models/Language & Arabic & Bengali & English & French & Spanish \\ [0.5ex] 
%  \midrule
%  XGLM(2.9B)-BLOOM(3B)  &  0.148 & 0.050 & 0.099 & 0.077 & 0.171 \\

%  \midrule
% XGLM(7.5B)-BLOOM(7B) & 0.185 & 0.058 & \textbf{0.005} & 
% \textbf{0.01} & \textbf{0.001} \\

% \bottomrule
% \end{tabular}
% }
% \caption{Testing fluency of the generations assuming different models of the same size generate relatively the same amount of incohesive texts. P-values determined by a permutation test with two samples of 300 observations per sample each coming from the pair of matching model sizes. P-values less than the chosen threshold, p < 0.05, are in bold }
% \asim{ask ben to revise the caption}
% \label{table:4}
% \end{table}

\subsection{Cohesiveness of Generation} 

%The difference in the quality of the generated samples is seen mostly in terms of the cohesiveness of the generated texts. 
As seen in Table \ref{table:incohesiveness_percentages}, BLOOM(7.1B) generates significantly more cohesive texts than XGLM(7.5B) for English, French, and Spanish with p-values under 5\%, based on a permutation-based statistical test.

% However, this trend does not follow in the case of Arabic. BLOOM, for Arabic, generates notably more incohesive texts than XGLM. BLOOM(3B) generates the most incohesive texts with a percentage of 13.3{\%} for Arabic. This high percentage is due to the incohesive dialog-based samples that BLOOM tends to output. We notice that while the sub-parts of a generated dialog are cohesive on their own, the dialog  as a whole is not natural. For Bengali, BLOOM(3B) surprisingly generates the least incohesive texts with a percentage of 4.3\%, which is lower than those of BLOOM(7.1B), XGLM(7.5B), and XGLM(2.9B).  

Interestingly, the results in Table \ref{table:incohesiveness_percentages} also show that a larger model does not necessarily lead to more cohesive generations. For example, BLOOM(3B) generates more cohesive texts than BLOOM(7.1B) for Bengali and English. XGLM(2.9B) also generates more cohesive texts than XGLM(7.5B) for English, French, and Spanish. We note that we are only evaluating cohesiveness in a binary way (cohesive vs. incohesive) and are not judging the quality of the predictions beyond that. 

Besides, the percentage of incohesive texts is noticeably higher for some languages than others for both BLOOM and XGLM. For example, the highest percentage of incohesive texts in the case of Bengali, English, and Spanish is less than or equal to 10\%, while that percentage is higher in the case of Arabic and French. 

\begin{table}
\resizebox{\columnwidth}{!}{
\begin{tabular}{c c c c c c} 
 \toprule
 \textbf{Model/Languag}e & \textbf{Arabic} & \textbf{Bengali} & \textbf{English} & \textbf{French} & \textbf{Spanish} \\ [0.5ex] 
 \midrule

  XGLM(2.9B) & \cellcolor{lightgreen}92\% & \cellcolor{pink}-3\%  & \cellcolor{lightgreen}14\% & \cellcolor{lightgreen}41\% & \cellcolor{lightgreen}58\%  \\
  
 BLOOM(3B) & \cellcolor{lightgreen}100\% & \cellcolor{pink}-6\% & \cellcolor{pink}-6\% & \cellcolor{pink}\textbf{-1\%*}  & \cellcolor{lightgreen}79\% \\

 \midrule

XGLM(7.5B) & \cellcolor{lightgreen}
\textbf{83\%*} & \cellcolor{lightgreen}33\% & \cellcolor{lightgreen}8\% & \cellcolor{lightgreen}32\% & \cellcolor{lightgreen}45\%  \\

BLOOM(7.1B) & \cellcolor{lightgreen}100\% & \cellcolor{pink}\textbf{-3\%*} & \cellcolor{pink}-13\% & \cellcolor{lightgreen}14\%& \cellcolor{lightgreen}67\%  \\ 
 
\bottomrule
\end{tabular}
}
\caption{Differences between formal and informal sample percentages of 400 samples per language (100 samples per model) sampled with neutral prompts. A green color indicates a bias toward formal generations and a pink color indicates a bias toward informal generations. Bolded values show that the corresponding model is significantly better according to a permutation-based statistical test with a p-value of 5\% or less.
}
\label{table:formality_bias}
\end{table}

\begin{table*}
\resizebox{\linewidth}{!}{
\begin{tabular}{c c c c c c} 
 \toprule
 \makecell{\multirow{2.3}{*}{\textbf{\textbf{Model/Language}}}} & \textbf{Arabic} & \textbf{Bengali} & \textbf{English} & \textbf{French} & \textbf{Spanish} \\ [0.5ex] 
& \textbf{F$\rightarrow$F\% / I$\rightarrow$I\%} & 
 \textbf{F$\rightarrow$F \% / I$\rightarrow$I\%} &
 \textbf{F$\rightarrow$F\% / I$\rightarrow$I\%} &
  \textbf{F$\rightarrow$F\% / I$\rightarrow$I\%} &
 \textbf{F$\rightarrow$F\% / I$\rightarrow$I\%}\\ [0.5ex] 

 \midrule
XGLM(2.9B) & 89.4\% / 61.1\% & 79.8\% / \textbf{100.0\%}* & 34.0\% / 94.0\% & 26.7\% / 59.5\% & 85.9\% / 80.2\% \\

BLOOM(3B) & 94.2\% / 55.1\% & 83.7\% / 87.1\% & 29.2\% / 91.7\% & 32.0\% / \textbf{82.0\%*} & 77.8\% / 90.4\% \\  

 \midrule

XGLM(7.5B) & 88.6\% / \textbf{76.7\%*} & 75.5\% / 98.8\% & 34.4\% / 84.7\% & \textbf{54.0\%*} / 75.6\% & 86.9\% / 75.8\% \\

 BLOOM(7.1B) & 93.5\% / 51.1\% & 74.0\% / 91.9\% & 27.6\% / \textbf{94.0\%*} & 25.8\% / 66.7\% & 83.8\% / \textbf{96.8\%*} \\

\bottomrule
\end{tabular}
}
\caption{Formality preservation samples' percentages for \textbf{Formal / Informal} prompts (800 prompts per language: 400 formal and 400 informal). Each sample is annotated as either formal, informal, or incohesive and the percentages are calculated without incohesive text counts. Bolded values show that the corresponding model is significantly better according to a permutation-based statistical test with a p-value of 5\% or less.
}
\label{table:formality_preservation}
\end{table*}

\subsection{Formality-Level Bias} \label{formality-bias}

Neutral prompts, given to an assumingly unbiased model, should lead to equitable distributions of formal and informal generations with a difference close to zero between both generations. However, this is not the case here as we show in Table \ref{table:formality_bias}. In the case of Bengali, we see that XGLM(2.9B), BLOOM(3B) and BLOOM(7.1B) are almost neutral with small differences of -3\% -6\% and -3\%, respectively, showing bias toward informal generations. On the other hand, we see XGLM(7.5B), surprisingly, showing significantly more bias toward formal generations than BLOOM(7.1B) with a difference of 33\%. Upon qualitative analysis, we found that many of the generations of XGLM(7.5B) had Bengali religious Islamic text-like attributes that were considered formal during annotation and the usage of hashtags or emojis was also less than the smaller model for neutral prompts. % However other formality markers such as the usage of second-person pronoun and their agreement with verb form, exclamation marks, and presence of ellipsis were similar for both models. 

BLOOM, for French, continues to show less bias showing only a bias of 1\% toward informal generations in the case of BLOOM(3B) and 14\% towards formal generations in the case of BLOOM(7.1B). On the other hand, XGLM(2.9) shows significantly more bias than BLOOM(3B) toward formal generations with a difference of 41\%. For English, XGLM and BLOOM both show %close bias in terms of the percentages while each showing bias toward a different direction. 
a small bias (in terms of percentages) towards different directions. XGLM(2.9B) and  XGLM(7.5B) show bias towards formal generations by 14\% and 8\% respectively. However, BLOOM(3B) and BLOOM(7.1B) display bias towards informal generations by 6\% and 13\% respectively. After a careful review of the predictions, we find that French and English informal predictions of BLOOM are due to a large proportion of informal generated dialog. %Indeed, \ben{TODO}

BLOOM, this time for Spanish, shows extreme bias towards the formal generations with a difference of 79{\%} for BLOOM(3B) and 67{\%} for BLOOM(7.1B). On the other hand, XGLM exhibits less bias towards formal generations with a difference of 58{\%} for XGLM(2.9B) and  45{\%} for XGLM(7.5B). These values indicate that both models are influenced by formal sources. In fact, most of the generated sequences with short prompts have the style of news titles/contents and Wikipedia articles. 

A biased distribution of outputs could be reasoned by the data the model was trained on. As stated in BLOOM \citep{scao2022bloom}, the biggest part of the corpus for Arabic was the Arabic-focused Masader repository \citep{masader, masaderplus}, which is dominated by Modern Standard Arabic (MSA) that is considered formal according to our definition of formality in section \ref{sec:formality}. This explains the extreme bias BLOOM(3B) and BLOOM(7.1B) show towards formal generations with a bias of 100{\%}. XGLM(7.5B) similarly shows an extreme bias toward formal generations, but significantly less than BLOOM(7.1B) with a difference of 83\%.

In terms of model size, we notice that XGLM(2.9B) shows more bias towards formal or informal generations than XGLM(7.5) for all the languages except Bengali, which could indicate that the bigger the XGLM model's size, the less biased it is. On the other hand, this isn't the case for BLOOM as BLOOM(3B) is only expressing more bias for Bengali and Spanish, while BLOOM(7.1B) shows more bias for English and French. 

In summary, the models show moderate bias for some languages such as English and Bengali, except for XGLM(7.5B) in the case of Bengali, while also showing extreme bias for other languages such as Arabic, French, and Spanish. This difference might be caused by the fact that every language is present in the data with a different percentage and is coming from different sources as shown in Table~\ref{tab:blooom_xglm_details}. Overall, it is noticeable that the bias is mostly toward formal generations for all the models and for all the languages.

\subsection{Formality-Level Preservation} \label{formality-preserve}

In this experiment, we measure how well the formality level of a generation is the same as the formality level of the prompt (i.e. how well the model preserves the formality-level of the prompt). We find that the formality style of the prompts is preserved efficiently for some languages by some models while being almost ignored in some other cases. 

For Arabic, as we show in Table \ref{table:formality_preservation}, BLOOM(3B) and BLOOM(7.1B) preserve the formality style of 94.2{\%} and 93.5\%, respectively, of the samples when the given prompt is formal. However, BLOOM does not pay that much attention to the style of the informal prompts and preserves the style of only 55.1\% of the samples with BLOOM(3B) and 51.1\% of the samples with BLOOM(7.1B). This confirms our finding from section \ref{formality-bias} that showed that BLOOM is biased toward formality in Arabic. XGLM(7.5B), on the other hand, preserves the informal style of the prompts significantly better than BLOOM(7.5B) with a percentage of 76.7\%. 

XGLM(2.9B), for Bengali, preserves the style of the informal prompts of significantly more samples than  BLOOM(3B) with a percentage of 100\%. BLOOM pays attention to the informal style of the prompts as well, unlike the case for Arabic, and preserves the style of 87.1\% of the samples generated with BLOOM(3B) and 91.9\% of the samples generated with BLOOM(7.1B). % XGLM and BLOOM preserve the formal style as well preserving the formality style of at least 74.0\% of the samples. 

Both BLOOM and XGLM, this time for English, do not preserve the formal style of the prompts for more than 34.4\% of the samples for any model. However, they both preserve the informal style in at least 84.7\% of the generated samples with BLOOM(7.1B) preserving significantly more samples than XGLM(7.5B). A similar trend follows for French with both BLOOM and XGLM unable to preserve the formal style for more than 32.0\% of the samples in the case of XGLM(2.9B), BLOOM(3B) and BLOOM(7.1B). On the other hand, XGLM(7.5) preserves the formal style significantly better than BLOOM(7.1B) with a percentage of 54.0\%. And again the informal style is being preserved better with, specifically, BLOOM(3B) which preserves the style better than XGLM(2.9B) with a percentage of 82\%. 

The formal and informal styles in  Spanish are preserved consistently across the models to at least 77.8\% of the samples with formal prompts and at least 75.8\% with informal prompts with BLOOM(7.1B) preserving the style in significantly more samples than XGLM(7.5B). 

In terms of model size, we notice that the size of the model is not an indicator of how well the model can preserve the formality style. For example, BLOOM(3B) preserves the formal style better than BLOOM(7.1B) for all languages except Spanish. In summary, we see that the informal style is mostly preserved well for most languages except with BLOOM for Arabic. The formal style, on the other hand, is mostly preserved well for all languages except English and French. 

\subsection{General Statistics about Generations} \label{characteristics-generation}
We report in Table \ref{table:genera_stats_of_generation} general statistics about the generated texts of each model and language by formality level. Results show that BLOOM generates about twice longer texts as XGLM. In terms of the average number of sentences per generation, BLOOM, when the generation is informal, generates more and shorter sentences than when the generation is formal. Also, informal generations tend to have emojis as expected, especially in the case of Bengali. Besides, informal generations tend to have more punctuation marks than formal ones. %which was not expected at all, however, this did not change the fact of them being informal. 
Finally, the results of the average number of new lines and the average number of ``-'', which are used to signal dialogues, support what we mentioned earlier about BLOOM's tendency to generate conversational text. 

\section{Discussion}

Formality bias when present in multilingual models, which are increasingly popular nowadays, can lead to undesirable outcomes. For example, using \textit{"please"} is common among North American English native speakers in requests, even among close friends, while in Arabic, it could be considered awkward, if not rude, in conversations among close friends  \cite{hovy-yang-2021-importance}. A usage example of language models is solving downstream tasks using prompting techniques for zero-shot learning, such as \cite{zhong2021adapting}'s work on question-answering. Prompting has also been used to utilize large language models for conversational chatbots such as ChatGPT \cite{ouyang2022training}. As prompting is becoming popular, we must understand that prompting a model that exhibits formality bias could be a barrier to getting the expected output. Furthermore, depending on the application, formality bias could even lead to sometimes unwanted misunderstandings \citep{hershcovich-etal-2022-challenges} and conflicts if the models, for example, are not able to generate text in the formality style of the users' expectations. 

Controlling LLMs generations has been taken into consideration in recent work, such as \cite{ouyang2022training}, which fine-tuned a language model \cite{NEURIPS2020_1457c0d6} intending to align the model with the intent of the users using reinforcement learning from human feedback (RLHF) \cite{christiano2017deep, stiennon2020learning}. Future work could analyze the impact of RLHF on the formality distributions present in language models. Furthermore, our work focused only on two pre-trained models with up to 7B parameters. The same analysis could be conducted for larger models such as GPT-3 and BLOOM(175B). Finally, the increase in the number of multilingual language models calls for more work on the bias analysis of multilingual language models.

\section{Conclusion}
In conclusion, we analyzed the formality level of the generations of two large-scale generative language models, XGLM and BLOOM, ranging from 2B parameters to 7B parameters. We first observed the cohesiveness of the predictions. We found that BLOOM(7.1B) predicts significantly more cohesive text than XGLM(7.5B) for English, French, and Spanish. Second, we showed that, across all five languages, both models tend to generate formal text when prompted neutrally. Finally, we found that the formality of the prompt highly impacts both models. In most cases, they generate the same style as the prompt, with slight differences between the models depending on the language. Our analysis is based on the annotations of 1,200 generations in Arabic, Bengali, English, French, and Spanish. We release them with this paper opening future avenues for modeling the formality of generative multilingual language models. 
%Formality is essential to how we perceive language. Our work quantifies the formality biases of two popular multilingual generative models across five languages. %Deploying generative language models successfully across a wide range of languages calls for Our work paves the way for a better understanding of how multilingual language models adapts to the culture and language. 
\section{Acknowledgment}

We thank the Fatima Fellowship\footnote{cf. \url{https://www.fatimafellowship.com/}} and Hugging Face for organizing and sponsoring the Fatima Research Fellowship program. 

% \section{Conclusion}

% Entries for the entire Anthology, followed by custom entries
\bibliography{anthology,custom}
\bibliographystyle{acl_natbib}
\newpage
\appendix

\section{Generation parameters}
\label{sec:appendix2}

\begin{table}[h]
\footnotesize
\centering
\begin{tabular}{ cccc }
 \toprule
  & \textbf{Top-k} & \textbf{Top-p} & \textbf{Temperature} \\
  \midrule
 Arabic & 50 & 0.95 & 1\\
 Bengali & 50 & 0.95 & 1\\
 English & 50 & 0.95 & 1\\ 
 French & 50 & 1 & 0.8\\
 Spanish & 50 & 1 & 0.8\\ 
 
 \bottomrule
\end{tabular}
\caption{Language-specific generation parameters for both models
}
\label{tab:gen_parameters}
\end{table}

Table \ref{tab:gen_parameters} shows details of the language-specific generation parameters we used for both BLOOM and XGLM.

\section{Descriptive statistics of the generations}

General statistics of the generations are in Table \ref{table:genera_stats_of_generation} reported per language for each model and generation label pair. The table contains the following statistics: the average length of the generation, the average number of sentences in a generation, the average length of the sentences, the average number of emojis per generation, the average number of punctuation marks per generation, the average number of new lines per generation, and finally, the average number of the dialogue mark/dash (-) per generation. 

\label{sec:appendix3}

\section{XGLM and BLOOM training corpora}
\label{sec:appendix1}

We show in Table \ref{tab:blooom_xglm_details} details of the languages used in our analysis in the training corpus of BLOOM and XGLM. 

\begin{table*}[t]
\footnotesize
\centering
\begin{tabular}{ cp{2.7cm}p{2.7cm}p{4cm}p{4cm} }
 \toprule
  & \multicolumn{2}{c}{\textbf{Corpus Size (GiB)}} & \multicolumn{2}{c}{\textbf{Data source domains}}\\
  \cmidrule{2-3}
  \cmidrule{4-5}
  & \makecell{\textbf{XGLM}} & \makecell{\textbf{BLOOM}} & \makecell{\textbf{XGLM}} & \makecell{\textbf{BLOOM}}\\
  \midrule
 \textbf{ar} & \makecell{64.34* (0.88\%)\\$\sim$66 upsampled} & \makecell{69.71 (4.34\%)} & \makecell{Web} & Web, news, books, subtitles, Wikipedia, wikisources\\
 \textbf{bn} & \makecell{11.19* (0.15\%) \\ $\sim$50 upsampled} & \makecell{17.32 (1.15\%)} & \makecell{Web} & Web, Wikipedia, Wikisource, open-source NLP datasets\\
 \textbf{en} & \makecell{3,324.45 (45.66\%)} & \makecell{451.64 (30.04\%)} & \makecell{Web} & Papers, Web, patents, books, subtitles, forums, Wikipedia, news \\ 
 \textbf{fr} & \makecell{303.76 (4.17\%)} & \makecell{193.94 (12.90\%)} & \makecell{Web} & Web, scholarly documents from all academic fields (HAL), Wikisource, Wikipedia, subtitles\\
 \textbf{es} & \makecell{363.83 (4.99\%)} & \makecell{163.07 (10.84\%)} & \makecell{Web} & Web, subtitles, Wikipedia, news, magazines\\ 
 \bottomrule
\end{tabular}
\centering
\caption{XGLM \citep{lin2021few} and BLOOM \citep{scao2022bloom} quantity and data sources in pre-training corpus.
%*XGLM authors do not mention the exact amount of GiB after the upsampling process.
% \ben{check the upsampling effect}
% \\
% \ben{choose between listing domains or dataset names}
}
\label{tab:blooom_xglm_details}
\end{table*}

\begin{table*}[h]
\centering

\resizebox{\linewidth}{!}{

\begin{tabular}{cccccccc}

\toprule 
    % Arabic Results
    \multicolumn{8}{c}{Arabic} \\
    \hline
    Prompt/\\Statistic 
    
    & Avg. Length & Avg. \# of sentences & Avg. Length of sentences
    & Avg. \# of emojis & Avg. \# of punctuation marks & Avg. \# of new lines & Avg. \# of dialogue mark(-)\\
    \midrule
    $XGLM(2.9B)-Informal$ & 175.074& 1.338& 130.593& 0.000& 6.794& 0.000& 0.000 \\
    $XGLM(2.9B)-Formal$ & 246.696& 1.686& 145.895& 0.010& 4.755& 0.000& 0.000 \\
    \midrule
    $BLOOM(3B)-Informal$ & 446.444& 2.037& 218.664& 0.000& 5.463& 2.185& 0.019 \\
    $BLOOM(3B)-Formal$ & 495.403& 3.583& 137.523& 0.005& 6.820& 3.626& 0.345 \\
    \midrule
    $XGLM(7.5B)-Informal$ & 187.345& 1.471& 127.023& 0.000& 9.391& 0.000& 0.000 \\
    $XGLM(7.5B)-Formal$ & 244.610 & 1.620& 150.581& 0.000 & 3.299 & 0.000 & 0.000 \\
    \midrule
    $BLOOM(7.1B)-Informal$ & 441.538 & 2.692 & 163.357 & 0.000 & 5.404 & 2.019 & 0.058 \\
    $BLOOM(7.1B)-Formal$ & 506.123 & 3.185 & 158.176 & 0.000 & 5.905 & 2.645 & 0.104 \\
    \bottomrule

    \multicolumn{8}{c}{Bengali} \\
    \hline
    Prompt/\\Statistic 
    
    & Avg. Length & Avg. \# of sent. per gen. & Avg. Length of sent.
    & Avg. \# of emojis per gen. & Avg. \# of punctuation marks per gen. & Avg. \# of new lines per gen. & Avg. \# of dialogue mark(-)\\
    \midrule
    $XGLM(2.9B)-Informal$ & 151.734& 1.552& 97.431& 0.357& 17.487& 0.000& 0.000 \\
    $XGLM(2.9B)-Formal$ & 164.149& 1.256& 130.467& 0.000& 3.091& 0.000& 0.000 \\
    \midrule
    $BLOOM(3B)-Informal$ & 413.338& 2.128& 193.667& 0.128& 11.047& 1.561& 0.020 \\
    $BLOOM(3B)-Formal$ & 384.252& 1.338& 286.909& 0.014& 5.518& 1.288& 0.007 \\
    \midrule
    $XGLM(7.5B)-Informal$ & 167.110& 1.728& 96.294& 0.360& 15.507& 0.000& 0.000 \\
    $XGLM(7.5B)-Formal$ & 152.767& 1.248& 122.199& 0.008& 2.880& 0.000& 0.000 \\
    \midrule
    $BLOOM(7.1B)-Informal$ & 419.400& 1.845& 226.829& 0.187& 13.484& 2.058& 0.155 \\
    $BLOOM(7.1B)-Formal$ & 418.500& 1.198& 349.046& 0.000& 5.063& 1.127& 0.008 \\
    \bottomrule

    \multicolumn{8}{c}{English} \\
    \hline
    Prompt/\\Statistic 
    
    & Avg. Length & Avg. \# of sent. per gen. & Avg. Length of sent.
    & Avg. \# of emojis per gen. & Avg. \# of punctuation marks per gen. & Avg. \# of new lines per gen. & Avg. \# of dialogue mark(-)\\
    \midrule
    $XGLM(2.9B)-Informal$ & 225.720& 3.332& 67.047& 0.005& 7.518& 0.000& 0.000 \\
    $XGLM(2.9B)-Formal$ & 261.529& 3.103& 83.544& 0.000& 5.943& 0.000& 0.000 \\
    \midrule
    $BLOOM(3B)-Informal$ & 584.288& 10.236& 56.152& 0.014& 19.803& 6.159& 0.620 \\
    $BLOOM(3B)-Formal$ & 646.354& 6.829& 93.727& 0.000& 12.537& 2.159& 0.000 \\
    \midrule
    $XGLM(7.5B)-Informal$ & 241.613& 3.497& 68.359& 0.022& 8.680& 0.000& 0.006 \\
    $XGLM(7.5B)-Formal$ & 281.921& 3.371& 82.887& 0.000& 6.000& 0.000& 0.000 \\
    \midrule
    $BLOOM(7.1B)-Informal$ & 575.236& 10.718& 52.733& 0.005& 22.278& 7.204& 1.324 \\
    $BLOOM(7.1B)-Formal$ & 639.466& 6.808& 93.020& 0.027& 14.123& 2.959& 0.110 \\
    \bottomrule

    \multicolumn{8}{c}{French} \\
    \hline
    Prompt/\\Statistic 
    
    & Avg. Length & Avg. \# of sent. per gen. & Avg. Length of sent.
    & Avg. \# of emojis per gen. & Avg. \# of punctuation marks per gen. & Avg. \# of new lines per gen. & Avg. \# of dialogue mark(-)\\
    \midrule
    $XGLM(2.9B)-Informal$ & 207.861& 2.723& 75.713& 0.058& 8.927& 0.000& 0.000 \\
    $XGLM(2.9B)-Formal$ & 231.435& 2.652& 86.646& 0.000& 6.861& 0.000& 0.000 \\
    \midrule
    $BLOOM(3B)-Informal$ & 621.216& 11.869& 51.417& 0.006& 25.562& 8.273& 1.051 \\
    $BLOOM(3B)-Formal$ & 612.727& 6.125& 99.197& 0.000& 13.909& 2.205& 0.034 \\
    \midrule
    $XGLM(7.5B)-Informal$ & 208.567& 2.850& 72.525& 0.047& 9.323& 0.000& 0.000 \\
    $XGLM(7.5B)-Formal$ & 235.277& 2.891& 80.735& 0.000& 7.403& 0.000& 0.000 \\
    \midrule
    $BLOOM(7.1B)-Informal$ & 588.804& 13.375& 43.091& 0.006& 29.667& 10.583& 2.607 \\
    $BLOOM(7.1B)-Formal$ & 637.415& 6.500& 97.216& 0.000& 15.679& 2.425& 0.066 \\
    \bottomrule

    \multicolumn{8}{c}{Spanish} \\
    \hline
    Prompt/\\Statistic 
    
    & Avg. Length & Avg. \# of sent. per gen. & Avg. Length of sent.
    & Avg. \# of emojis per gen. & Avg. \# of punctuation marks per gen. & Avg. \# of new lines per gen. & Avg. \# of dialogue mark (-)\\
    \midrule
    $XGLM(2.9B)-Informal$ & 222.789& 2.798& 78.974& 0.028& 9.514& 0.000& 0.000 \\
    $XGLM(2.9B)-Formal$ & 249.69& 2.228& 111.517& 0.000& 6.123& 0.000& 0.000 \\
    \midrule
    $BLOOM(3B)-Informal$ & 553.59& 9.291& 58.689& 0.000& 21.000& 5.427& 0.846 \\
    $BLOOM(3B)-Formal$ & 613.827& 4.532& 134.672& 0.000& 12.012& 1.734& 0.012 \\
    \midrule
    $XGLM(7.5B)-Informal$ & 225.454& 2.981& 74.957& 0.019& 9.870& 0.000& 0.000 \\
    $XGLM(7.5B)-Formal$ & 248.728& 2.32& 106.663& 0.006& 6.254& 0.000& 0.000 \\
    \midrule
    $BLOOM(7.1B)-Informal$ & 530.218& 8.589& 60.846& 0.000& 20.565& 5.435& 1.331 \\
    $BLOOM(7.1B)-Formal$ & 640.643& 4.661& 136.668& 0.000& 12.393& 1.655& 0.012 \\
    \bottomrule

\end{tabular}}
\caption{The average statistics per formal and informal generation of sequence length, number of sentences, length of sentences, number of emojis, number of punctuation marks, number of new lines, and number of the dialogue mark (``-'').}
\label{table:genera_stats_of_generation}
\end{table*}

\section{Formality Distribution}

\begin{figure}[h]
    \centering
    \resizebox{\columnwidth}{!}{
    \includegraphics[height=6.8cm]{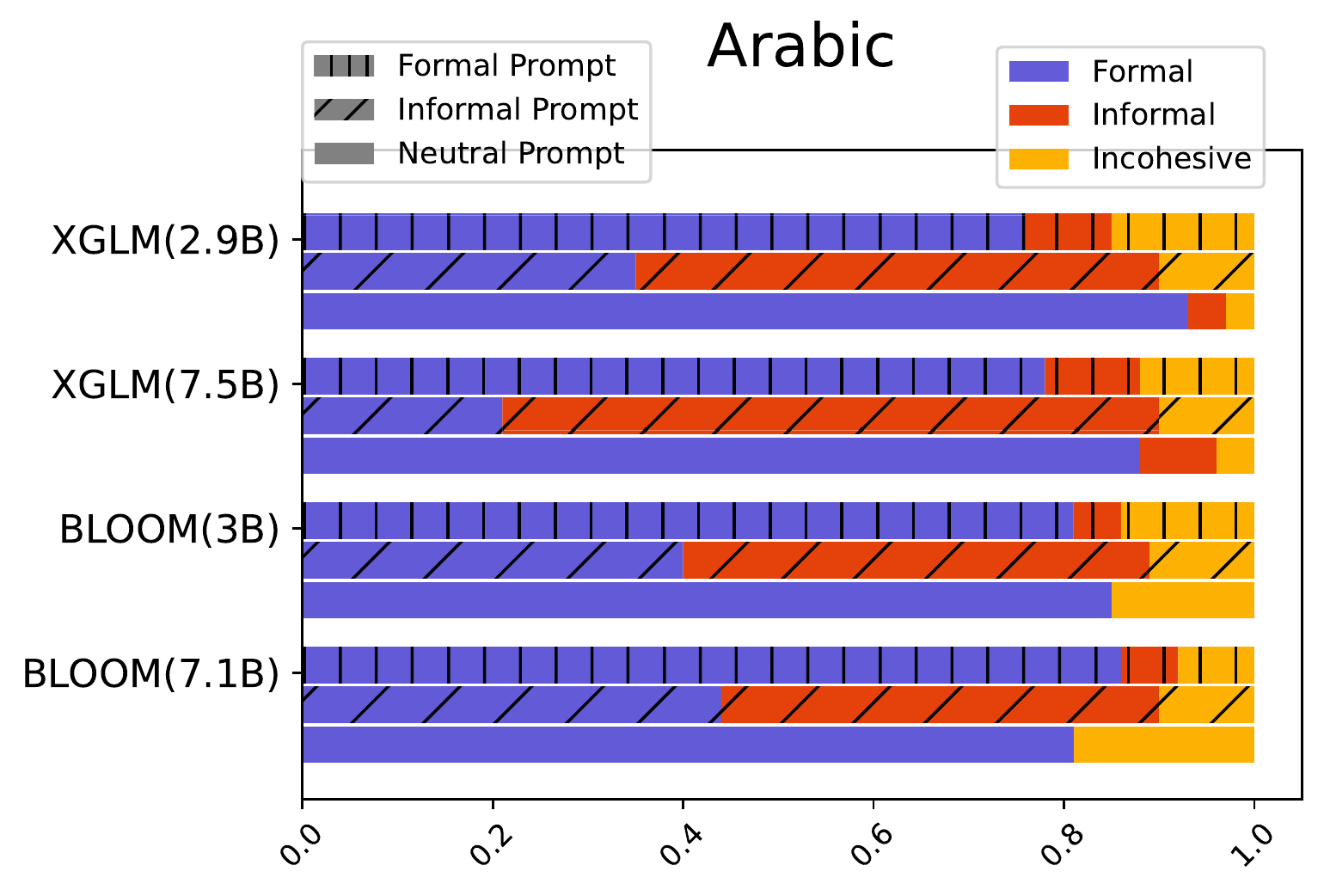}
    }
    \caption{Plot of the distribution of the generations for Arabic, for each prompt type, according to their labeled categories: Formal, Informal, and Incohesive. Each bar in the plot represents 100 generations.}
    \label{fig:arabic_fig}
\end{figure}

\begin{figure}[h]
    \centering
    \resizebox{\columnwidth}{!}{
    \includegraphics[height=6.8cm]{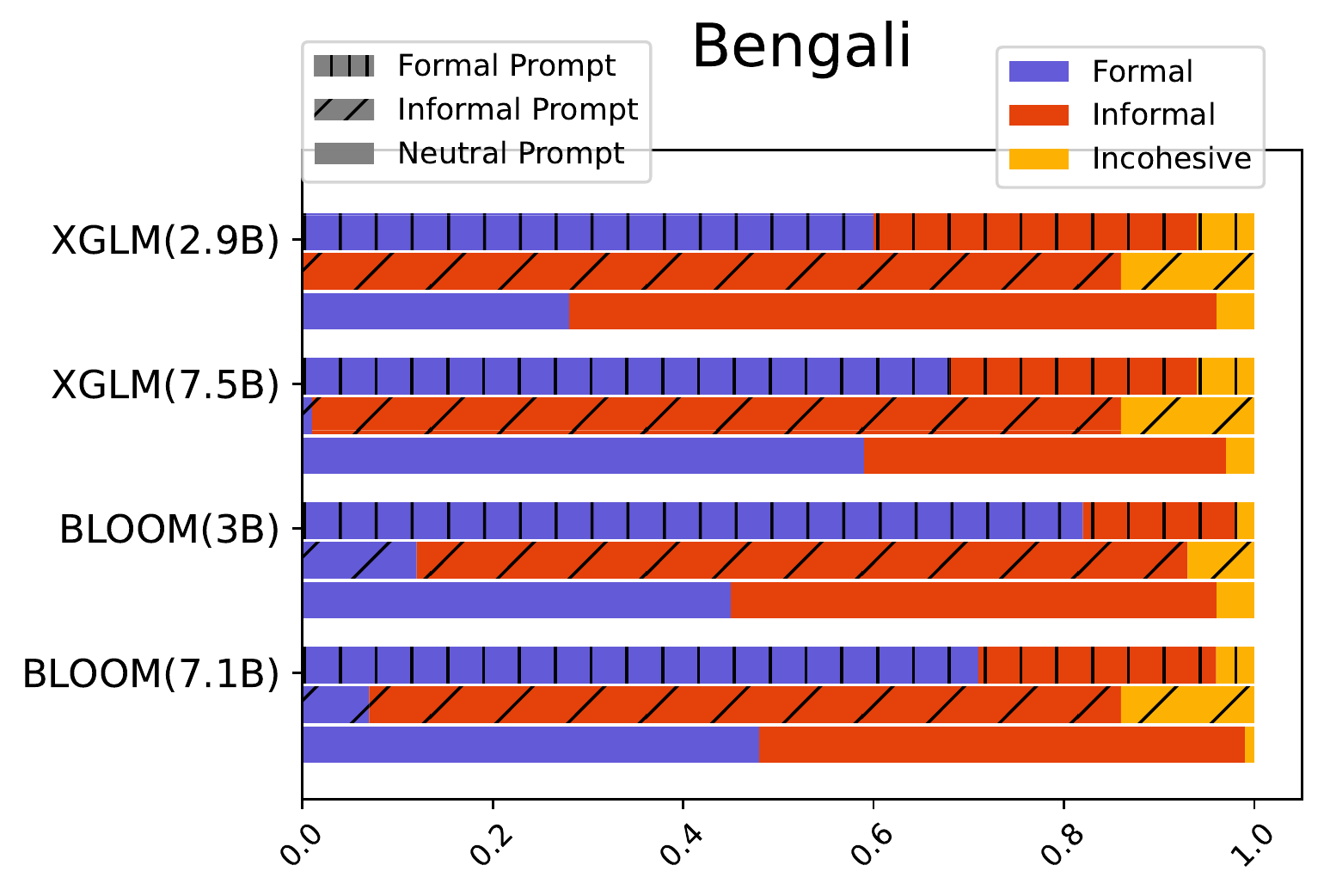}
    }
    \caption{Plot of the distribution of the generations for Bengali, for each prompt type, according to their labeled categories: Formal, Informal, and Incohesive. Each bar in the plot represents 100 generations.}
    \label{fig:bengali_fig}
\end{figure}

\begin{figure}[h]
    \centering
    \resizebox{\columnwidth}{!}{
    \includegraphics[height=6.8cm]{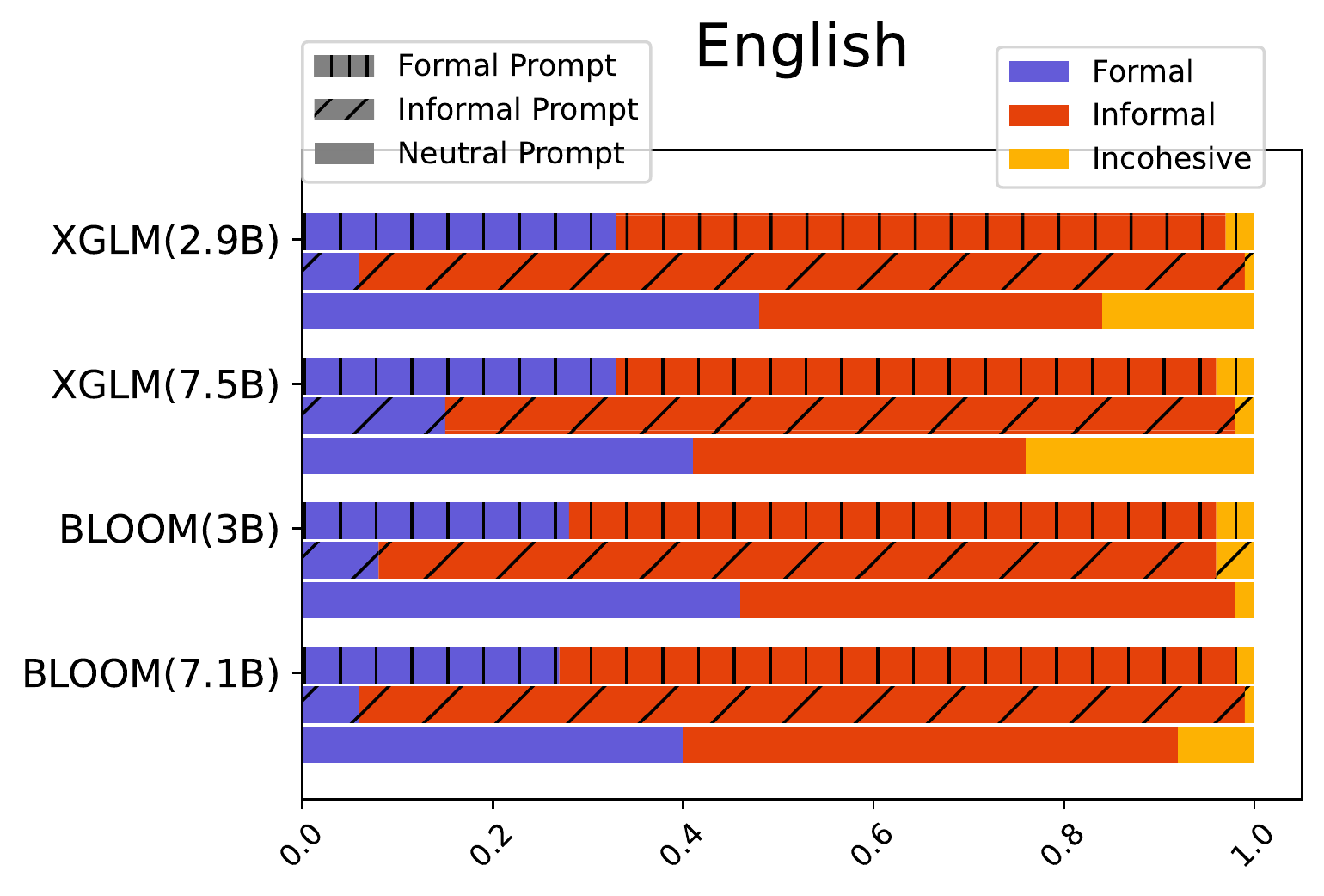}
    }
    \caption{Plot of the distribution of the generations for English, for each prompt type, according to their labeled categories: Formal, Informal, and Incohesive. Each bar in the plot represents 100 generations.}
    \label{fig:english_fig}
\end{figure}

\begin{figure}
    \centering
    \resizebox{\columnwidth}{!}{
    \includegraphics[height=6.8cm]{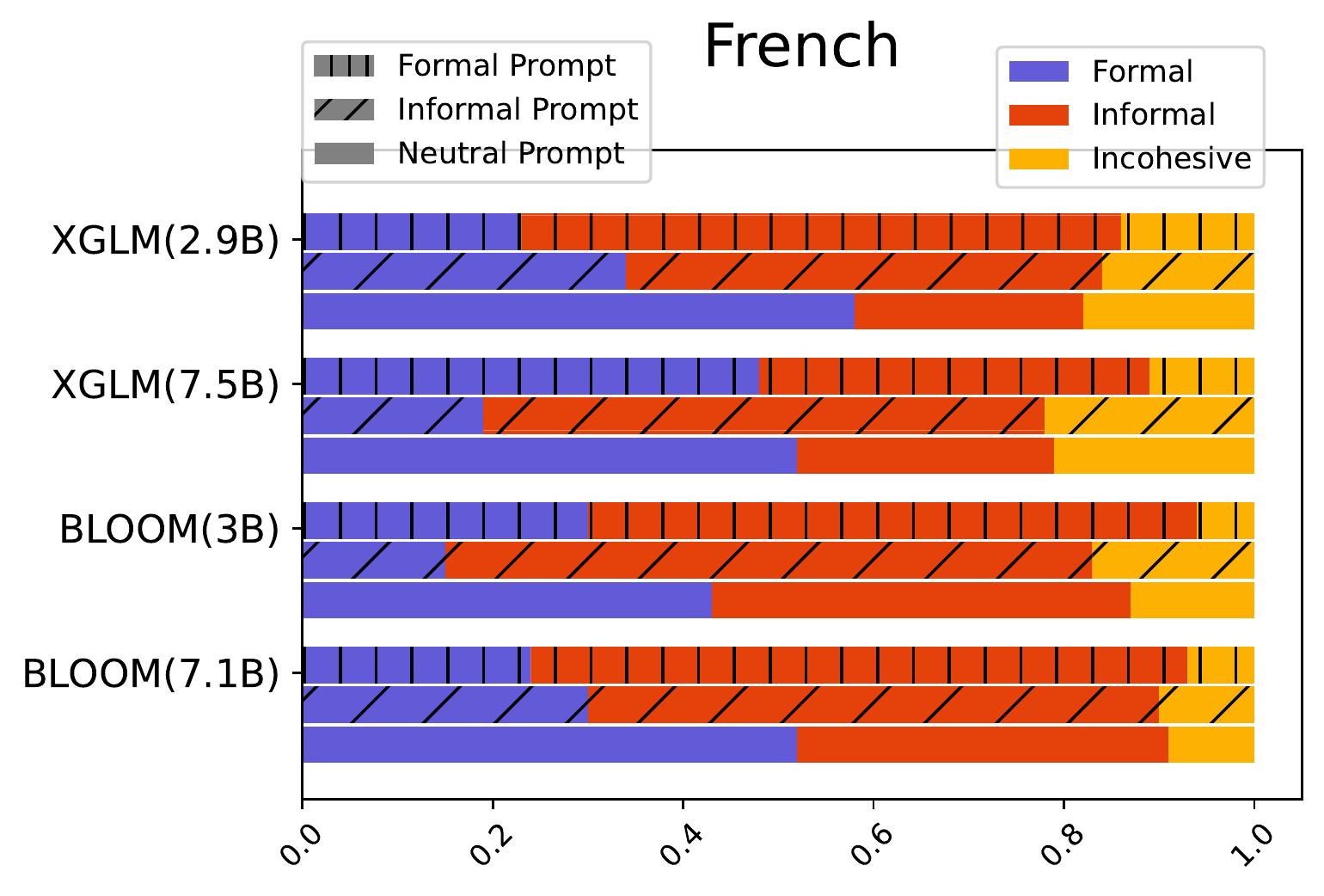}
    }
    
    \caption{Plot of the distribution of the generations for French, for each prompt type, according to their labeled categories: Formal, Informal, and Incohesive. Each bar in the plot represents 100 generations.}
    \label{fig:french_fig}
\end{figure}

\begin{figure}
    \centering
    \resizebox{\columnwidth}{!}{
    \includegraphics[height=6.8cm]{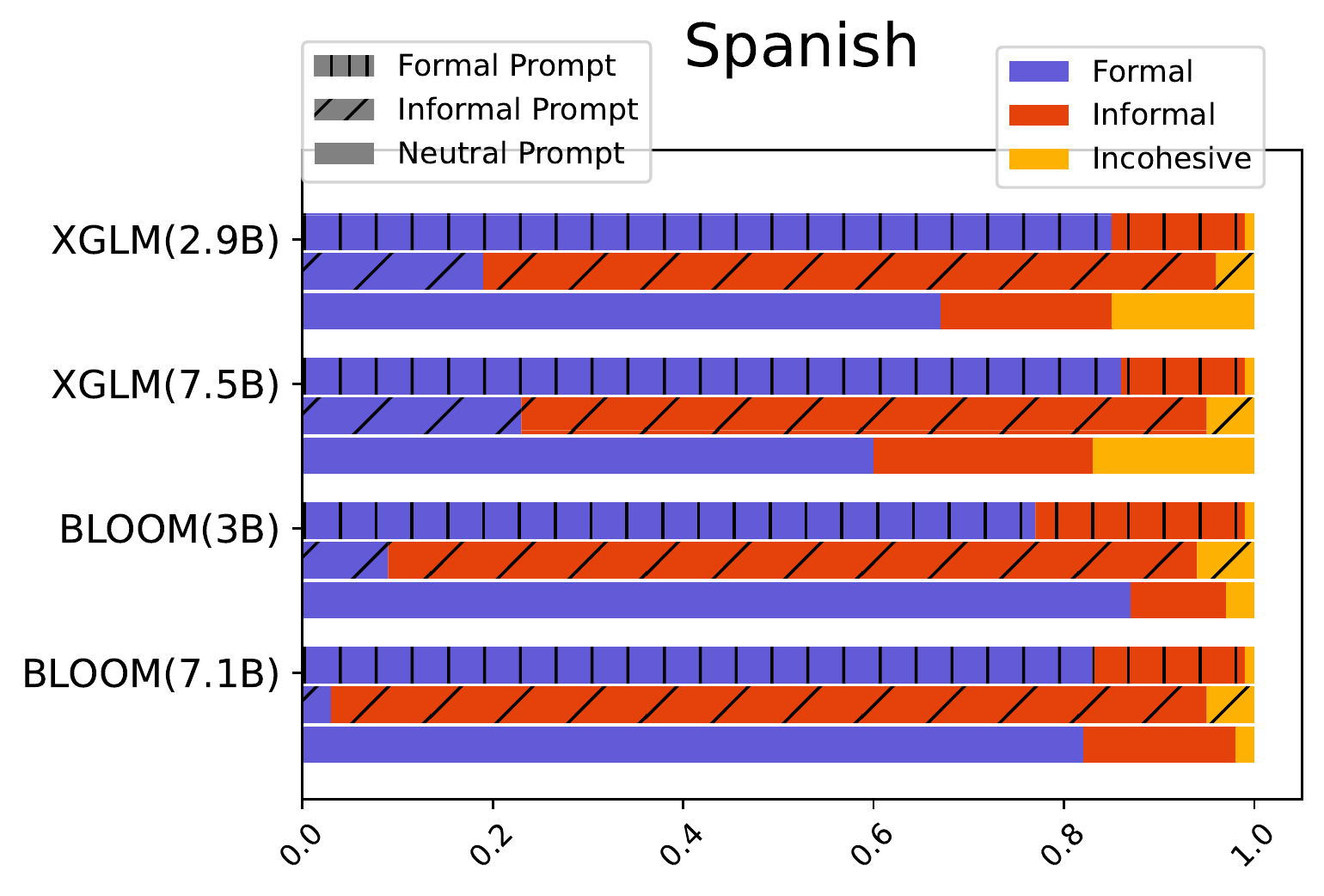}
    }
    
    \caption{Plot of the distribution of the generations for Spanish, for each prompt type, according to their labeled categories: Formal, Informal, and Incohesive. Each bar in the plot represents 100 generations.}
    \label{fig:spanish_fig}
\end{figure}

We visualize the annotated data for each language to help in seeing an overview of all the results. Each language is represented by a plot, see Figures \ref{fig:arabic_fig}, \ref{fig:bengali_fig}, \ref{fig:english_fig}, \ref{fig:french_fig}, and \ref{fig:spanish_fig}, with 12 bars with 3 bars corresponding to each model representing the 3 prompts types: formal informal, and neutral. Each bar in the plot represents 100 texts generated with the corresponding model when prompted with the corresponding prompt type. The colors in each bar represent the 3 possible annotations: formal, informal, and incohesive. 

\label{sec:appendix4}

\end{document}